\documentclass[lettersize,journal]{IEEEtran}
\usepackage{amsmath,amsfonts}
\usepackage{algorithmic}
\usepackage{algorithm}
\usepackage{array}
\usepackage[caption=false,font=normalsize,labelfont=sf,textfont=sf]{subfig}
\usepackage{textcomp}
\usepackage{stfloats}
\usepackage{url}
\usepackage{verbatim}
\usepackage{graphicx}
\usepackage{adjustbox}
\usepackage{cite}
\usepackage{amsmath}
\usepackage{amssymb}
\usepackage{graphicx}
\usepackage{multirow}
\usepackage{array}
\usepackage{caption}
\usepackage{booktabs}
\begin{document}

\title{TLD: A Vehicle Tail Light signal Dataset and Benchmark}

\author{Jinhao Chai, Shiyi Mu, Shugong Xu}

\markboth{}%
{Shell \MakeLowercase{\textit{et al.}}: A Sample Article Using IEEEtran.cls for IEEE Journals}


\maketitle

\begin{abstract}
Understanding other drivers' intentions is crucial for safe driving. The role of taillights in conveying these intentions is underemphasized in current autonomous driving systems. Accurately identifying taillight signals is essential for predicting vehicle behavior and preventing collisions. Open-source taillight datasets are scarce, often small and inconsistently annotated. To address this gap, we introduce a new large-scale taillight dataset called TLD. Sourced globally, our dataset covers diverse traffic scenarios. To our knowledge, TLD is the first dataset to separately annotate brake lights and turn signals in real driving scenarios. We collected 17.78 hours of driving videos from the internet. This dataset consists of 152k labeled image frames sampled at a rate of 2 Hz, along with 1.5 million unlabeled frames interspersed throughout. Additionally, we have developed a two-stage vehicle light detection model consisting of two primary modules: a vehicle detector and a taillight classifier. Initially, YOLOv10 and DeepSORT captured consecutive vehicle images over time. Subsequently, the two classifiers work simultaneously to determine the states of the brake lights and turn signals. A post-processing procedure is then used to eliminate noise caused by misidentifications and provide the taillight states of the vehicle within a given time frame. Our method shows exceptional performance on our dataset, establishing a benchmark for vehicle taillight detection. The dataset is available at https://huggingface.co/datasets/ChaiJohn/TLD/tree/main
\end{abstract}


%%%%%%%%%%%%%%%%%%%%%%%%%%%%%%%%%%%%%%%%%%%第一章%%%%%%%%%%%%%%%%%%%%%%%%%%%%%%%%%%%%%%%%%%%%

% 插入图片corner-case
\begin{figure}[h!]  % 'h!' 表示尽可能将图片放置在当前位置
    \centering
    \adjustbox{max width=\textwidth}{\includegraphics[width=1.0\linewidth]{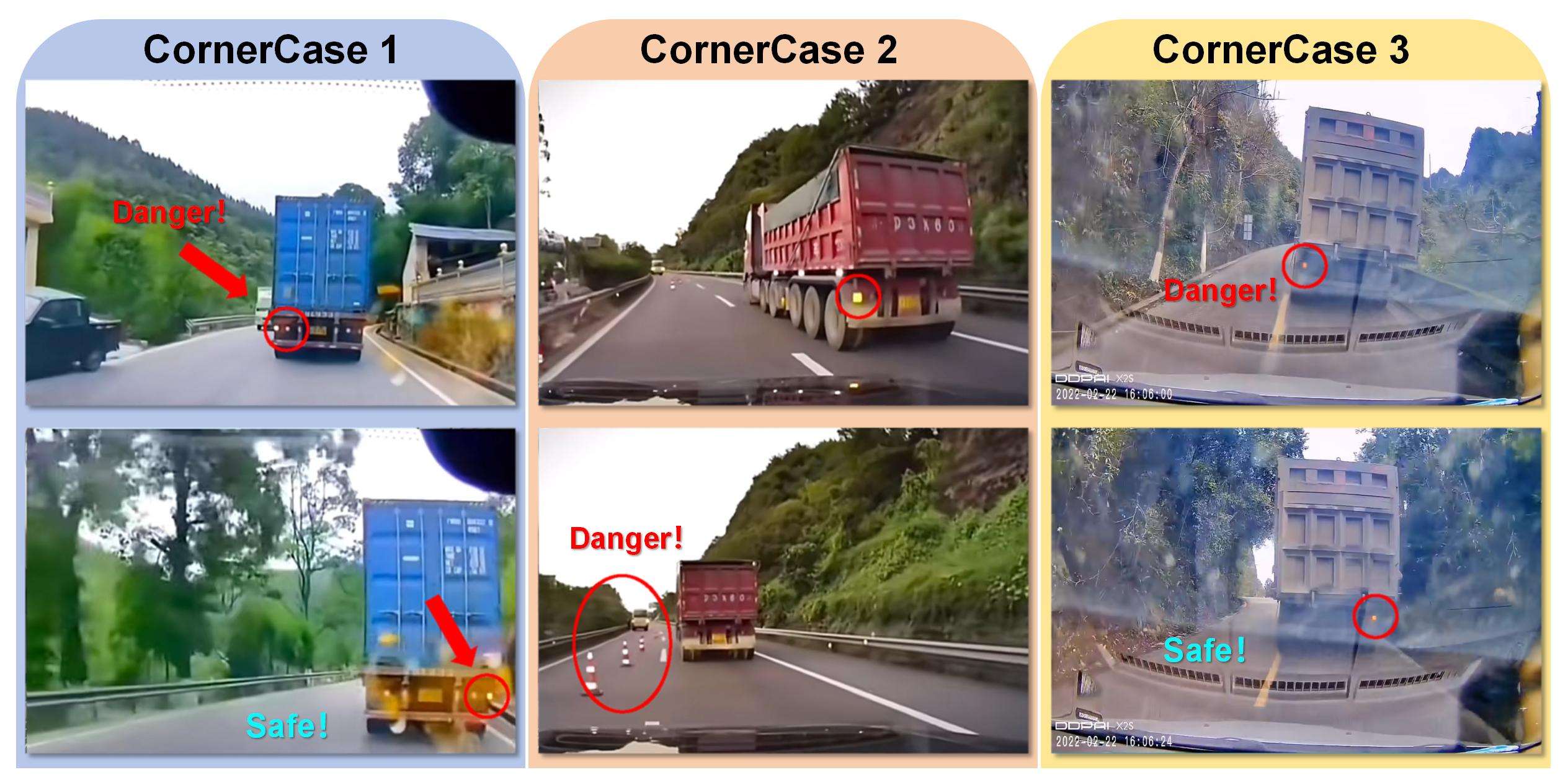}}  % 使用 adjustbox 进行自动缩放
\caption {Three corner cases where turn signal information is used for interaction to avoid danger. \textbf{Corner case 1 and 3:} The truck driver turns on the left turn signal to indicate to the following vehicle that there is an oncoming vehicle in the opposite lane, making it unsafe to overtake. When it is safe to overtake, the truck driver turns on the right turn signal to indicate that the following vehicle can overtake.
\textbf{Corner case 2:}The truck driver uses the left turn signal to indicate that the lane ahead is narrowing or there is an obstacle.}
    \label{fig:corner-case}  % 设置标签以便在文中引用
\end{figure}

% 插入图片show-dataset
\begin{figure*}[t]  % 't' 表示将图片放置在页面顶部
    \centering
    \adjustbox{max width=\textwidth}{\includegraphics[width=1.9\linewidth]{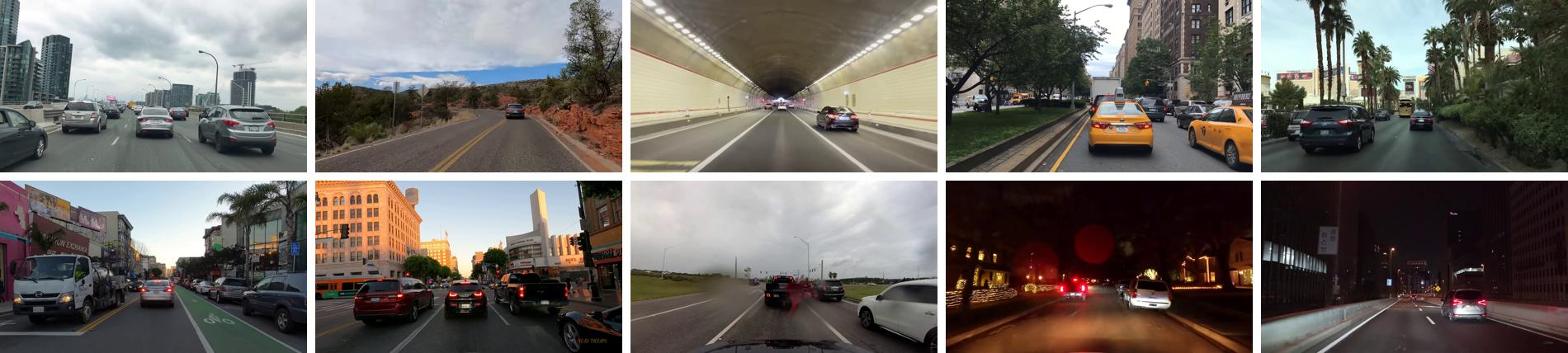}}  % 使用 adjustbox 进行自动缩放
    \caption{The driving scenarios in the \textbf{TLD }cover different times of the day (day/night), various weather conditions (sunny, rainy), and diverse settings (congested urban areas, highways, rural areas).}
    \label{fig:keshihua}  % 设置标签以便在文中引用
\end{figure*}

% 插入图片difficult
\begin{figure}[h!]  % 'h!' 表示尽可能将图片放置在当前位置
    \centering
    \adjustbox{max width=\textwidth}{\includegraphics[width=1.0\linewidth]{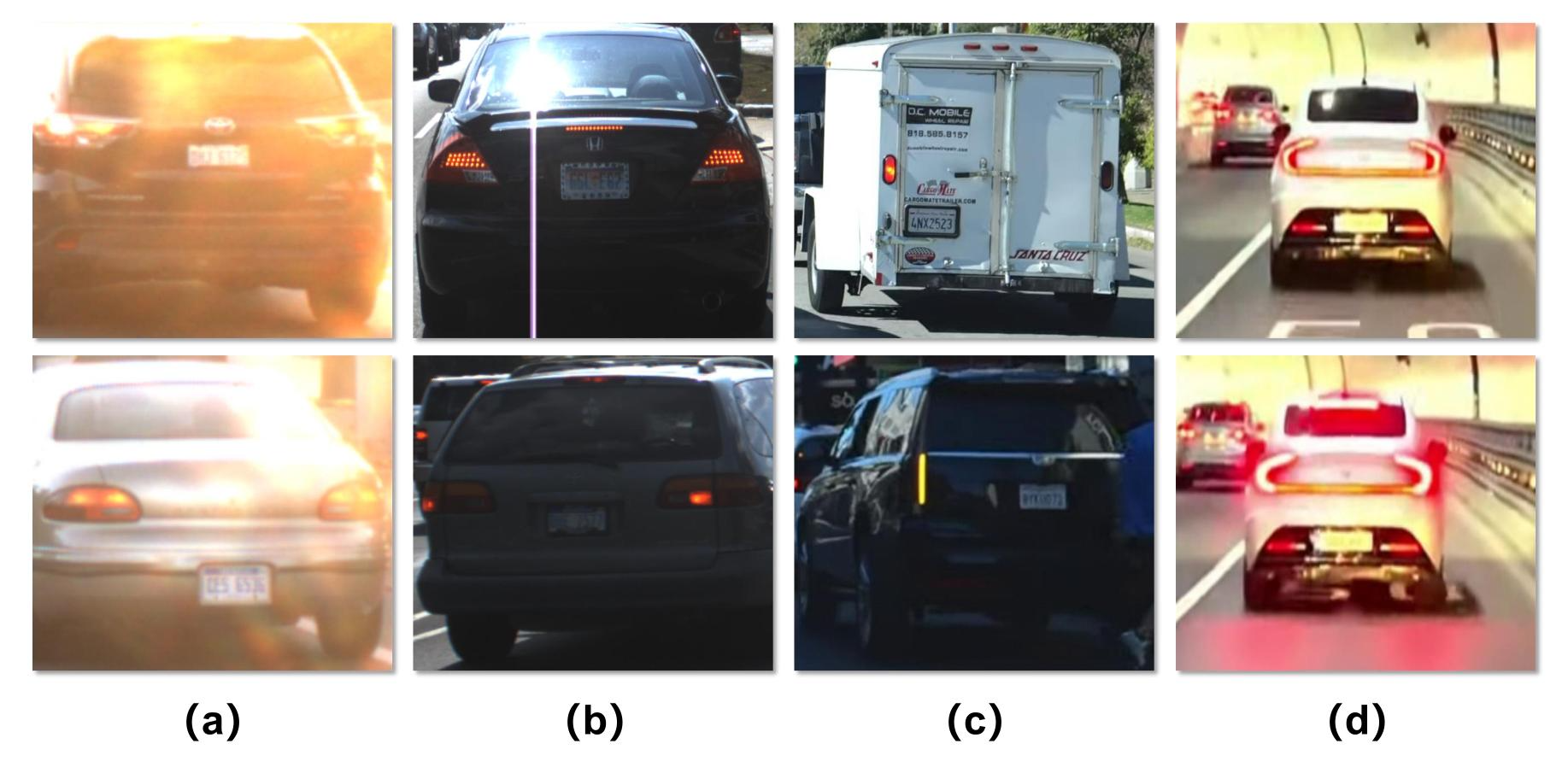}}  % 使用 adjustbox 进行自动缩放
\caption {Various lighting conditions, diverse shapes of tail lights, differing arrangements of lights between day and night, challenges with temporal integration, and insufficient resolution all contribute to making the detection of tail light status a particularly challenging task. }
    \label{fig:difficult}  % 设置标签以便在文中引用
\end{figure}

\section{Introduction}
\IEEEPARstart{I}{n} RECENT years,  with the continuous development and maturation of autonomous driving technology, an increasing number of vehicles equipped with advanced driver-assistance systems (ADAS) have appeared on the roads. In densely populated urban traffic scenarios, quickly and accurately perceiving the behavioral intentions of surrounding vehicles—and making safer, more intelligent decisions based on this perception—has become a primary focus for many autonomous driving system researchers. In real-world driving, vehicle taillight signals, which play a crucial role in communicating driving intentions among human drivers, can be seen as a visual language used to indicate forthcoming actions, such as turning or lane-changing, to other road users. Therefore, enabling autonomous driving systems to interpret this vehicle light language is vital for a better understanding of driving intentions.

Unlike traditional methods of inferring driving intentions, which typically involve estimating the next move of other vehicles based on their direction and speed, these approaches can sometimes lead to erroneous judgments. More critically, such methods often introduce a delay in understanding driving intentions. For example, when a vehicle in the left lane intends to change lanes into the ego vehicle's lane, an aggressive lane change might cause the system to fail in predicting the intention, potentially triggering emergency braking (AEB) or even causing a collision. However, by incorporating the recognition of vehicle taillights, the system gains a predictive element in judging other vehicles' driving intentions. This is because vehicle taillight signals are designed to pre-announce a driver's next action to surrounding vehicles, with turn signals usually activated before the actual maneuver.

Furthermore, recent studies have shown that clear reasoning about the long-term goals \cite{r1,r2,r3} and short-term intentions \cite{r4,r5,r6} of other vehicles in a traffic scene can significantly improve the accuracy of trajectory predictions, directly benefiting downstream tasks in autonomous driving systems. Additionally, this type of optical visual signal can be considered a simple form of vehicle-to-vehicle (V2V) communication. In Figure \ref{fig:corner-case}, we provide examples where taillight signals play an indispensable role in various corner cases. For instance, on mountain roads or highways, truck drivers generally have a broader field of view compared to smaller cars, allowing them to detect potential hazards earlier. In such scenarios, the rear vehicle, due to visual obstructions, may miss certain areas of perception. The lead vehicle can use its taillight signals to indirectly fill in these perceptual blind spots, effectively helping the rear vehicle avoid potential collision risks.

Based on this, we believe that detecting and recognizing vehicle taillights can enable autonomous driving systems to better understand the driving intentions and interaction signals of surrounding vehicles, thus achieving safer and more reliable driving.

However, in practice, current autonomous driving systems often lack mature solutions for vehicle taillight detection, and the information provided by other vehicles' lights is not fully utilized. This could be because perceiving taillight signals presents certain challenges. Based on our analysis, we have identified the following challenges in recognizing taillight states:

\begin{itemize}
\item {\bf Varying Lighting Conditions.} During the day, the red taillight cover on vehicles may reflect sunlight, making it appear as though the lights are on. At night, various light sources, such as streetlights and oncoming headlights, can interfere with taillight detection. Additionally, certain lighting conditions can introduce imaging noise, including strong reflections, halos, and shadows.
\end{itemize}

\begin{itemize}
\item {\bf Occlusions from Random Observation Angles.} In congested traffic scenarios, such as waiting at a red light, the random nature of observation angles can lead to partial occlusions between vehicles, making it difficult to accurately judge taillight states.

\end{itemize}

\begin{itemize}
\item {\bf Non-uniform Taillight Shapes and light forms.} The lack of unified standards among car manufacturers results in significant differences in taillight shapes across vehicles, including cars, vans, trucks, and buses. Moreover, some modern taillight designs, like strip lights that illuminate sequentially, pose additional challenges for detection.

\end{itemize}

\begin{itemize}
\item {\bf Inconsistent Taillight States Between Day and Night.} Vehicles generally keep their lights off during the day but turn on the side lights at night. Since side lights are close to the brake lights, detecting brake lights becomes more challenging at night, making the high-mounted brake light a better detection choice.
\end{itemize}

\begin{itemize}
\item {\bf Temporal Sequence Issues.} Turn signals typically blink to indicate activation, so determining the state of turn signals requires considering both the current frame and previous states in the time sequence. The overall state change in the time sequence ultimately determines the turn signal status.
\end{itemize}

In addressing these challenges, machine learning techniques have proven highly effective for pattern recognition \cite{r7,r8}, especially given the need for large amounts of data. However, in the field of vehicle light recognition, this presents a limitation. There is an urgent need for a large-scale taillight dataset that encompasses various lighting conditions, weather scenarios, viewing angles, and vehicle types to train and evaluate deep learning models for taillight detection. To meet this need, we introduce TLD, a new large-scale vehicle light detection dataset. TLD not only covers diverse traffic scenarios under different global weather and lighting conditions but also includes a sufficient number of challenging samples. Additionally, we propose a two-stage vehicle light detection model as a baseline. Experimental results demonstrate that our model achieves commendable performance on the dataset.

In conclusion, our work makes two main contributions. First, we introduce TLD (TailLight Dataset), the first publicly available large-scale vehicle light detection dataset. This dataset includes 152,690 images with annotated taillight states, featuring decoupled annotations for brake lights, turn signals, and hazard lights. The comprehensive taillight state annotations and diverse real images in TLD provide a solid foundation for improving the detection and recognition of automotive taillight signals. This can assist ADAS systems in better understanding the driving intentions of surrounding vehicles, thereby benefiting downstream tasks for safer and more intelligent planning and decision-making. We believe that our dataset will support further research in vehicle light detection and driving intention prediction within the community.Second, we establish a new baseline on our dataset using a simple yet effective method for recognizing taillight states in various scenarios. Experimental results indicate that our method not only effectively detects and recognizes taillight states of surrounding vehicles but also demonstrates robustness across different times (day and night), weather conditions (sunny/rainy), and locations (urban areas, highways, tunnels, rural areas, etc.).

%%%%%%%%%%%%%%%%%%%%%%%%%%%%%%%%%%%%%%%%%%%第二章%%%%%%%%%%%%%%%%%%%%%%%%%%%%%%%%%%%%%%%%%%%%

\section{RELATED WORKS}
In this section, we review some of the existing work in the field of vehicle light detection, which can generally be categorized into two approaches: image processing-based methods and deep learning-based methods. Since taillight colors are predominantly red, image processing methods often employ heuristic approaches using various color spaces such as HSV, YCrCb, Lab*, or Y'UV to detect red light regions in the rear of vehicles. These color space transformation methods extract candidate taillight regions by thresholding channels containing red components after converting the input images to different color spaces, and remove surrounding noise using morphological operations \cite{5218254,5446402,28} or noise filtering \cite{14,21}. Additionally, since taillights are symmetrically aligned along the vehicle's center axis, many methods use symmetry checks \cite{20!,21,5446402} and aspect ratio tests \cite{5218254,21} to identify the most probable taillight regions and perform taillight status recognition. Liu \cite{kajabad2018detection} used RGB space and set color difference thresholds between adjacent frames to determine brake light operation. Some studies choose to process images in the Lab* color space, where L* represents brightness and a* and b* represent the color ranges for the red-green and yellow-blue axes, respectively. Chen et al.\cite{7247631} utilized the a* component in the Lab* color space for binarization of brake light detection. Nava et al. \cite{8916961} and Pirhonen\cite{pirhonen2022brake} further explored color feature-based and morphological operation-based detection methods in the Lab* color space.

Apart from color space conversion, some studies have employed specific image processing techniques such as frequency domain analysis, brightness thresholds, and symmetry tests to enhance detection accuracy and robustness. For instance, Jen et al. \cite{7991176} proposed a fast radial symmetry transform algorithm for daytime brake light detection. Cui et al. \cite{cui2015vision} developed a hierarchical framework using a deformable parts model to detect vehicles and applied clustering techniques to extract taillight candidate regions.

Recent advancements in deep learning have significantly advanced vehicle light detection. These methods train convolutional neural networks (CNNs) to automatically learn image features without manual feature design. Hsu et al.\cite{8317782} introduced a CNN-LSTM structure capable of learning spatiotemporal features of vehicle taillights from video sequences, which not only improved detection accuracy but also enhanced the model's adaptability to dynamic changes. Lee et al. \cite{8814278} further integrated attention mechanisms, focusing on key regions in images and critical time steps in sequences to substantially improve vehicle taillight recognition performance.

Some research has focused on taillight detection under specific conditions such as nighttime or adverse weather. For example, Duan-Yu Chen et al.\cite{6740840} proposed a nighttime turn signal detection method based on Nakagami-m distribution, using scattering modeling and reflectance analysis to identify turn signal directions. Almagambetov et al. \cite{6328045} introduced an algorithm utilizing Kalman filters and codebooks for automatic tracking of vehicle taillights, as well as detection of brake lights and turn signals, demonstrating robust performance under varying lighting and weather conditions.

Moreover, many studies focus on optimizing taillight detection performance through various technical approaches. For instance, O'Malley et al.\cite{5446402} used taillight detection to improve the accuracy of nighttime vehicle detection. Skodras et al.\cite{6208089} and Thammakaroon et al.\cite{5218254} enhanced taillight detection algorithms through morphological operations and brightness analysis. Guo et al. \cite{9257398} and Jeon et al. \cite{9789175} improved detection robustness using deep learning frameworks and multi-view information fusion, respectively.

The aforementioned methods primarily focus on single-frame image-based taillight detection. However, taillight recognition is closely related to the state and actions over time, as discussed in Chapter 1. Consequently, some methods incorporate time series analysis. Inspired by video action classification problems, researchers have attempted to apply common video action classification techniques for taillight state time series analysis, such as two-stream, CNN-LSTM, and 3D convolutional (C3D) networks. The two-stream method \cite{12,13} uses RGB frames and multi-frame dense optical flow fields as inputs, processed through two CNNs to handle spatial and temporal information. CNN-LSTM \cite{14!,15} extracts spatial features from each frame using CNNs and learns temporal features with LSTM. C3D \cite{16,17} extends 2D convolution with temporal domain convolution, processing both spatial and temporal information simultaneously.

In related research, probabilistic graphical models are often used to handle variable-length non-image data sequences. Probabilistic graphical models include Hidden Markov Models (HMM), Maximum Entropy Markov Models (MEMM), and Conditional Random Fields (CRF), and are widely applied in fields such as natural language processing, future prediction \cite{18,19}, sequence classification \cite{20,r21}, and sequence labeling \cite{r22,r23}. For example, Huang et al. \cite{r24} combined LSTM with CRF to address the long-term dependency issue in sequence labeling, with CRF establishing long-term correlations between sequences. These methods provide valuable insights for taillight recognition in time series analysis.

Despite the significant achievements of existing methods in vehicle light detection, challenges remain, particularly in generalization under varying lighting conditions and complex environments, as well as meeting real-time requirements. Future research needs to focus on enhancing model robustness, reducing computational resource consumption, and developing more precise and real-time detection algorithms. With ongoing technological advancements, we are confident that vehicle light detection technology will mature and provide strong support for the development of autonomous driving and intelligent transportation systems.

% 插入分布图
\begin{figure}[t]  % 'h!' 表示尽可能将图片放置在当前位置
    \centering
    \includegraphics[width=0.9\linewidth, height=0.26\textheight]{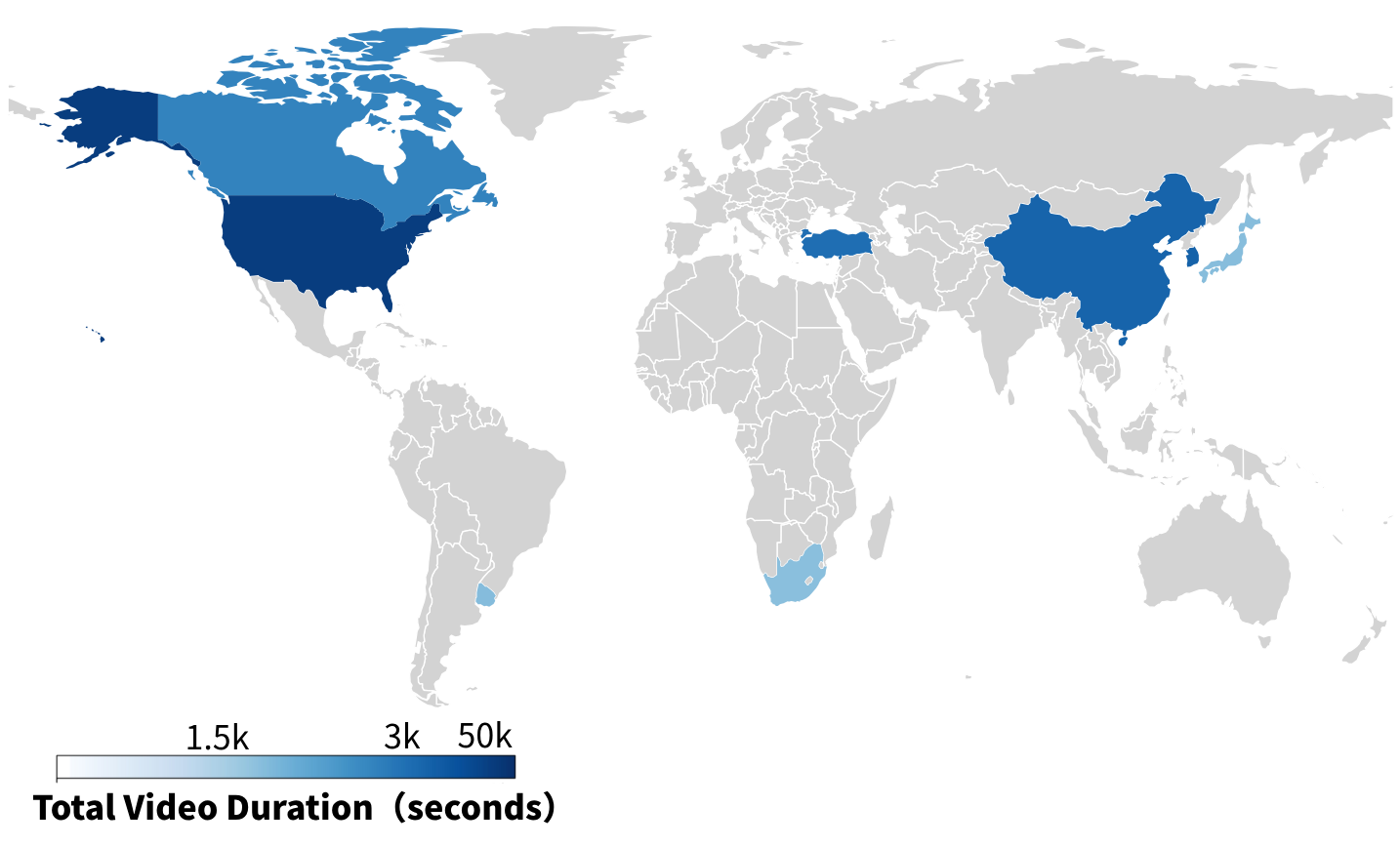}  % 使用 height 属性来调整图片的高度
    \caption{\textbf{Geographic distribution of TLD.} Our dataset covers ample driving scenarios around the world.}
    \label{fig:global}  % 设置标签以便在文中引用
\end{figure}

%%%%%%%%%%%%%%%%%%%%%%%%%%%%%%%%%%%%%%%%%%%第三章%%%%%%%%%%%%%%%%%%%%%%%%%%%%%%%%%%%%%%%%%%%%

\section{METHODOLOGY}

\subsection{Dataset}

\renewcommand{\arraystretch}{1.5} % 增加表格中的行间距

\begin{table*}[ht]
\centering
\caption{Comparison of datasets that feature vehicle taillight status.}
\label{tab:datasets}
\begin{tabular}{@{}>{\centering\arraybackslash}p{2.5cm} >{\centering\arraybackslash}p{0.6cm} >{\centering\arraybackslash}p{1.2cm} >{\centering\arraybackslash}p{1.5cm} >{\centering\arraybackslash}p{1.5cm} >{\centering\arraybackslash}p{1.2cm} >{\centering\arraybackslash}p{1.5cm} >{\centering\arraybackslash}p{1.0cm} >{\centering\arraybackslash}p{1cm} >{\centering\arraybackslash}p{0.8cm} >{\centering\arraybackslash}p{1.8cm}@{}}
\toprule
\textbf{Dataset} &
  \textbf{Year} &
  \textbf{Sequences} &
  \textbf{\begin{tabular}[c]{@{}c@{}}Frames\\ Annotated\end{tabular}} &
  \textbf{Instances} &
  \textbf{\begin{tabular}[c]{@{}c@{}}Turning\\ Signal\end{tabular}} &
  \textbf{\begin{tabular}[c]{@{}c@{}}LightState\\ Classes\end{tabular}} &
  \textbf{Day/Night} &
  \textbf{Rainy} &
  \textbf{Global} &
  \textbf{Task} \\ \midrule
VRSD\cite{8317782}            & 2017 & \checkmark & -      & 63637  & \checkmark & 8 & D   & $\times$ & $\times$ & Classification \\
VLS\cite{vls}             & 2022 & \checkmark & 4811   & -      & \checkmark & 4 & D/N & \checkmark & $\times$ & Detection      \\
Pirhonen et al.\cite{pirhonen2022brake} & 2022 & $\times$ & -      & 822    & $\times$ & 2 & D/N & $\times$ & $\times$ & Classification \\
Song et al.\cite{action}     & 2022 & \checkmark & -      & 287040 & \checkmark & 5 & D/N & \checkmark & $\times$ & Classification \\
LISA\cite{lisa}            & 2023 & $\times$ & -      & 40000  & \checkmark & 0 & D/N & $\times$ & $\times$ & Detection      \\
Oh et al.\cite{oh}       & 2023 & \checkmark & 11088  & -      & $\times$ & 2 & D/N & $\times$ & $\times$ & Detection      \\
Barshooi et al.\cite{ford} & 2023 & \checkmark & 3800   & -      & \checkmark & 4 & D/N & $\times$ & $\times$ & Detection      \\
Moon et al.\cite{moon}     & 2023 & \checkmark & 12000  & -      & $\times$ & 2 & D/N & \checkmark & $\times$ & Detection      \\ \midrule
Ours                    & 2024 & \checkmark & 152690 & 307509 & \checkmark & 8 & D/N & \checkmark & \checkmark & Detection      \\ \bottomrule
\end{tabular}
\end{table*}

Unfortunately, there are currently few publicly available large-scale datasets for vehicle taillight recognition in real driving scenarios. Much of the work in academia relies on self-recorded driving videos for research, and these individually collected datasets are typically not open-source. Moreover, due to the varying annotation requirements of different methods, these datasets lack general applicability. We conducted a systematic review of existing datasets in the field of taillight detection, analyzing their types and characteristics. A detailed comparison can be found in Table \ref{tab:datasets}.

% 插入柱状图
\begin{figure}[t]  % 'h!' 表示尽可能将图片放置在当前位置
    \centering
    \adjustbox{max width=\textwidth}{\includegraphics[width=1\linewidth]{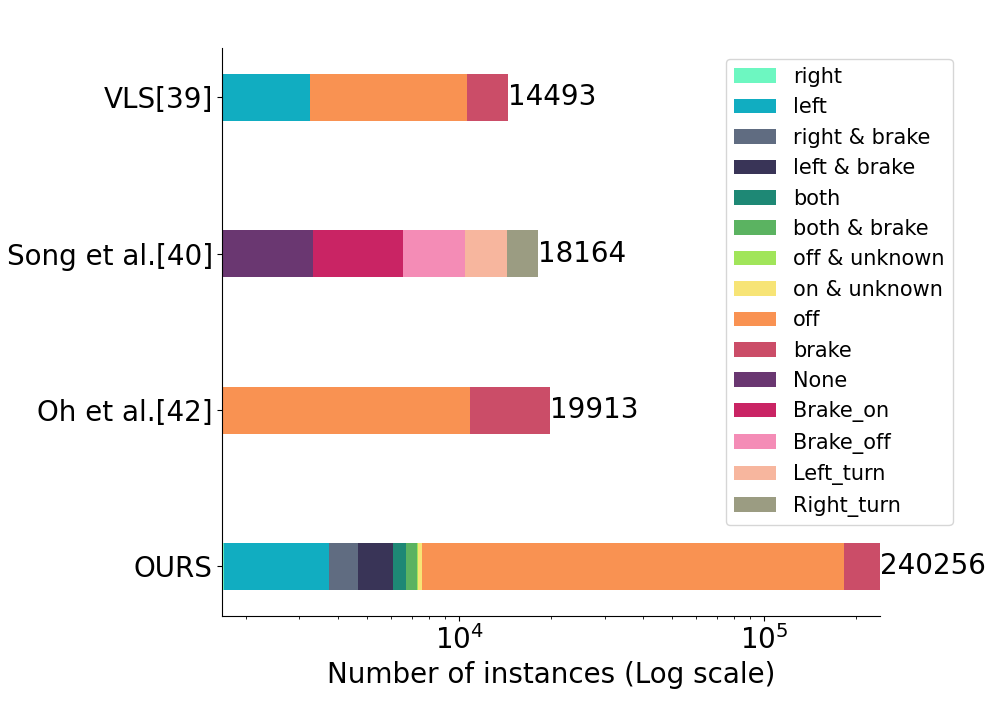}}  % 使用 adjustbox 进行自动缩放
\caption{\textbf{Comparison of the number and total counts of various categories between the TLD and other vehicle light datasets.} The horizontal axis is presented on a logarithmic scale. Our dataset significantly surpasses the others in both the number of categories and the overall quantity.}

    \label{fig:zhuzhuangtu}  % 设置标签以便在文中引用
\end{figure}

Among these taillight detection datasets, the LISA dataset \cite{lisa} focuses on localizing the taillight region using four corner coordinates to annotate the polygonal edges of the taillights. However, it does not include annotations for downstream taillight state recognition tasks. Some studies \cite{8317782,pirhonen2022brake,action,lisa} choose to annotate the taillight states of cropped vehicle images. While these datasets often contain a large number of cropped images, they lack comprehensive scene information, making it difficult to fully validate a taillight detection pipeline. For taillight state recognition tasks, depending on the specific focus of the dataset, some datasets \cite{pirhonen2022brake,oh,moon} only annotate brake light states, while others include both brake light and turn signal states. The latter annotations are more complex but provide richer and more comprehensive taillight state information.

Notably, for datasets that include both brake lights and turn signals, the academic community has largely classified taillight states into broad categories such as Brake-Off, Brake-On, Turn-Left, and Turn-Right. However, this classification method does not decouple turn signals from brake lights, which we believe is crucial in practical applications. In real traffic scenarios, it is common and reasonable for vehicles to have both brake lights and turn signals activated simultaneously, such as when a vehicle slows down to turn. The use of such coarse classification methods results in the loss of brake light information when turn signals are active, which is detrimental to accurately predicting other vehicles' driving intentions. Although Hsu et al. \cite{7247631} provided an eight-state taillight classification in their Vehicle Rear Signal Dataset, the dataset only includes cropped images of vehicles, rather than full driving scenes.

To address this, we present TLD (Tail Light Dataset), our proposed solution to the current pain point in taillight datasets: the lack of a large-scale dataset with decoupled annotations of brake lights and turn signals in extended real-world driving scenarios. TLD contains over 1.7 million images, including 152,690 annotated images and over 1.5 million unlabeled images, with a total of 307,509 annotated instances. The total duration of the driving videos is 17.78 hours. To our knowledge, TLD is the first large-scale public dataset to provide decoupled annotations of turn signals and brake lights in full-frame images from real-world driving scenarios.

Our dataset is entirely derived from real driving scenes, with annotated images sourced from high-quality driving videos on YouTube and supplemented with additional annotations from the LOKI dataset. Most of TLD’s images come from YouTube driving videos. We carefully selected 21 driving videos from the extensive high-quality YouTube driving video list compiled by the autonomous driving video dataset OpenDV-2K \cite{genad}, covering 15.53 hours of footage from diverse global driving scenes (Figure \ref{fig:global}). These videos span a wide range of common driving conditions, including day and night, various weather conditions, congested urban scenes, and suburban highway scenarios, offering significant scene diversity. To facilitate future research on time-series analysis, we extended the original videos to a 30Hz sampling rate, providing a substantial number of unlabeled images between manually annotated frames. This resulted in a total of 1.6 million frames. These unlabeled frames are valuable for semi-supervised learning, allowing models to improve generalization and performance by training on a combination of labeled and unlabeled data. They can also be used for generating pseudo-labels, which can be added to the training set to enhance model training. By combining labeled and unlabeled data, we can better capture variations and dynamic information in the video, thereby improving overall model robustness and generalization.

Another part of the dataset involves additional annotations on the LOKI dataset \cite{loki}, which was introduced by Honda Research Institute USA (HRI USA) in 2021. The LOKI dataset was collected in dense urban environments in downtown Tokyo and is divided into 644 scenes, each averaging 12.6 seconds in length. It already includes various agent intention labels (e.g., stopping, left turn, right turn, lane change to the left) and sensor information such as LiDAR. We manually annotated the taillight information for all visible vehicles in the LOKI dataset, with each driving scene downsampled to 5Hz for annotation, resulting in a total of 40,890 frames. We chose to annotate the LOKI dataset because we believe a dataset that includes both taillight states and agent intention labels will significantly contribute to further research on multi-agent interaction and the intrinsic connection between taillight information and driving intentions.

% 插入图片pipeline
\begin{figure*}[t]  % 't' 表示将图片放置在页面顶部
    \centering
    \adjustbox{max width=\textwidth}{\includegraphics[width=0.8\linewidth]{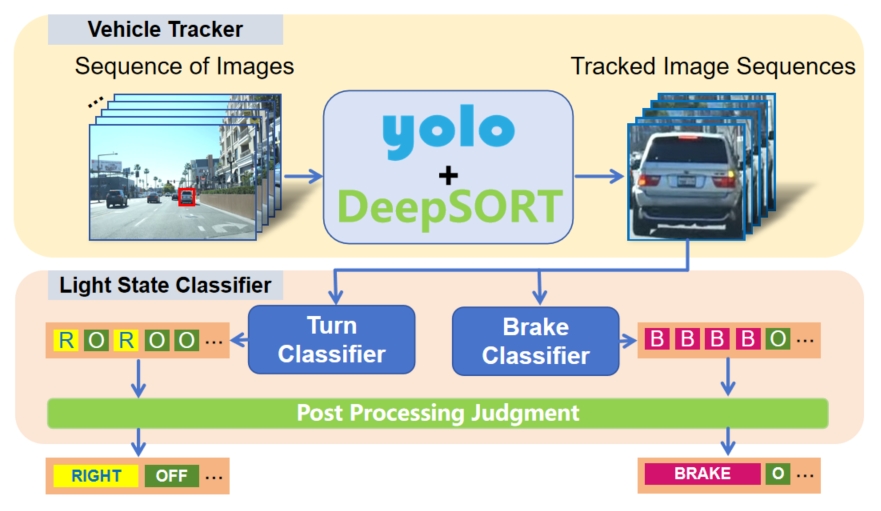}}  % 使用 adjustbox 进行自动缩放
    \caption{\textbf{Overview of the Taillight Detection Pipeline.}The video or image sequence is processed through YOLOv10 and DeepSORT for object tracking to obtain a sequence of cropped vehicle images. This sequence is then fed into parallel turn signal and brake light classifiers, frame by frame, to determine the status of the vehicle lights. Finally, the light status over a continuous period is determined using a post-processing module.}
    \label{fig:pipeline}  % 设置标签以便在文中引用
\end{figure*}
% 插入图片

We further divided TLD into two versions based on their sources, with more details provided in Section 4.1.

\subsection{Overview of the Taillight Detection Pipeline}

Our vehicle light detection method is both straightforward and efficient. It consists of two primary modules, as shown in Figure \ref{fig:pipeline}: the vehicle detection module and the tail light state classifier. The vehicle detection module's main task is to detect and track nearby vehicles in the captured driving scenes. These identified vehicle segments are then designated as regions of interest (ROI) and passed on to the tail light state classifier. The brake light and turn signal classifiers work in parallel to analyze the sequence of tracked vehicles. The resulting sequence of tail light states for each frame is then processed by a post-processing module, which filters out potential false detections and determines the turn signal's status over a given period.

\subsection{Vehicle Detection Module}

% 插入图片
\begin{figure}[h!]  % 'h!' 表示尽可能将图片放置在当前位置
    \centering
    \adjustbox{max width=\textwidth}{\includegraphics[width=1\linewidth]{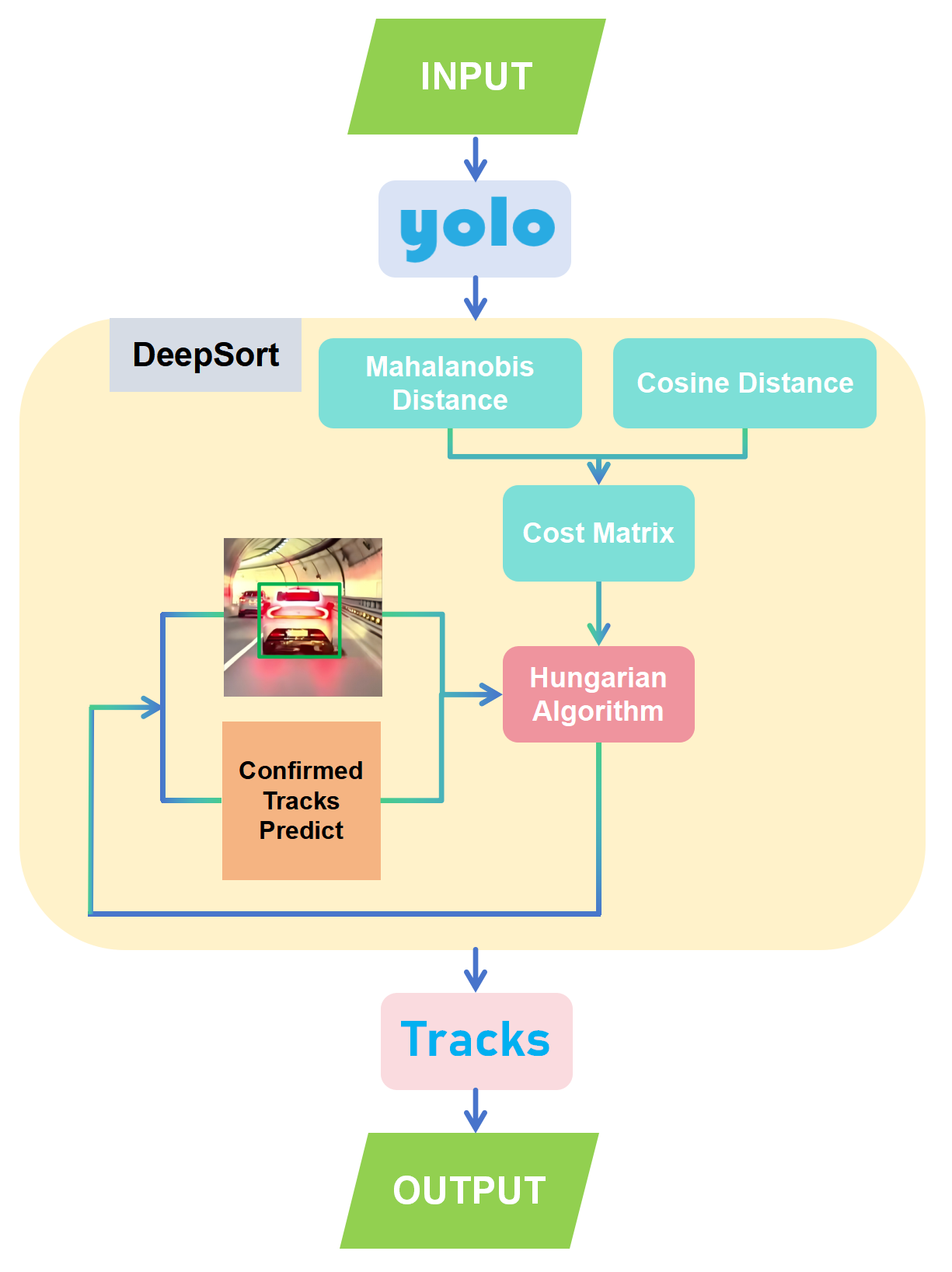}}  % 使用 adjustbox 进行自动缩放
    \caption{The operation process of YOLOv10-DeepSORT. The input is sent to YOLOv10 to detect the targets. Then, the ReID model requires a well-distinguishing feature embedding from the output of YOLOv10. The Kalman filter predicts the trajectory of each target. Finally, the outputs of YOLOv10 and the Kalman filter are sent to the Hungarian algorithm for matching in order to obtain the trajectory of the target.}
    \label{fig:pipeline2}  % 设置标签以便在文中引用
\end{figure}

The vehicle detection module aims to extract the Region of Interest (ROI) for the tail light state classifier. Its goal is to minimize interference from unrelated factors in the driving scene, such as traffic lights, street lights, and reflective barriers. To detect and track other vehicles, we use a combination of YOLOv10\cite{yolov10} and DeepSort\cite{deepsort}.

The YOLO (You Only Look Once) series models redefined object detection by mapping input images directly to bounding box coordinates and class probabilities in an end-to-end manner. This single-stage framework effectively overcomes the high computational costs associated with R-CNN and its subsequent methods, such as Fast R-CNN and Faster R-CNN. The YOLO series has become one of the most popular object detection methods in practical applications. This approach not only significantly enhances detection speed but also reduces reliance on computational resources while maintaining high accuracy, providing a clear advantage in real-time object detection tasks.

Considering the real-time requirements of due to its efficiency, excellent detection accuracy, and real-time capabilities.

For tail light detection, we chose YOLOv10 as the base model for vehicle detection. It is currently the best-performing model and we opted for the TINY version to reduce inference latency. YOLOv10 features a deeper convolutional neural network structure, incorporating the latest feature extraction techniques and efficient attention mechanisms. It also includes multi-scale feature fusion modules and adaptive anchor box generation strategies, which further improve detection accuracy and robustness. YOLOv10-TINY, the chosen model, consists of 16 convolutional layers, 5 max pooling layers, 1 upsampling layer, 1 attention mechanism layer, and 2 output layers. It uses a Feature Pyramid Network (FPN) strategy to predict bounding boxes at two different scales: 20×20 and 40×40. By optimizing the model structure and reducing the number of parameters, YOLOv10-TINY delivers satisfactory detection performance on embedded and mobile devices while maintaining high detection speed and accuracy.The vehicle tracking network's purpose is to continuously track detected vehicles' positions over time to ensure accurate capture of their dynamics. We employed the DeepSort algorithm (Simple Online and Realtime Tracking with a Deep Association Metric) for this module. Traditional SORT algorithms (Simple Online and Realtime Tracking) use Kalman filters to predict target positions in the future and the Hungarian algorithm for data association. However, SORT relies solely on target motion information, which can lead to errors in crowded scenes or when targets are occluded. DeepSort improves SORT by introducing a deep learning-based appearance feature extraction module. This module uses convolutional neural networks to extract feature vectors for each target, supplementing the Kalman filter and Hungarian algorithm to enhance data association accuracy and robustness.

Specifically, the vehicle tracker workflow involves the following main steps: First, the YOLOv10 model detects vehicles in video frames, generating target detection results called "detections." Each detection typically includes information about the target, such as bounding box coordinates and confidence scores. Next, a pre-trained deep convolutional neural network extracts feature vectors from each detected target region. For Confirmed Tracks Predict, the Kalman filter is applied to the next frame to estimate the new position and velocity of the tracks, which are then associated with the detections in the current frame. During the association step, the Mahalanobis distance (equation \ref{eq:mahalanobis}) between detection targets and tracking targets is computed, where \(z\) is the observed target state (detection target), \(\hat{x}\) is the predicted state of the tracking target, and \(S\) is the state covariance matrix. If the Mahalanobis distance is below a specified threshold, they are matched as the same target. However, since Mahalanobis distance can struggle with occlusion, DeepSort also employs cosine distance for appearance similarity. A ReID model extracts feature vectors for different objects, and a cosine distance cost function computes the similarity between the predicted and detected objects. If the bounding boxes are close and the features are similar, they are matched as the same target. Once data association is complete, DeepSort updates the state of each tracked target, including position, velocity, and feature vectors. For newly appearing vehicles, DeepSort initializes a new Kalman filter and begins tracking them. For vehicles that remain unmatched for a long time, the system removes the corresponding tracker to reduce computational resource consumption.

\begin{equation}
D_{\text{Mahalanobis}} = (z - \hat{x})^\top S^{-1} (z - \hat{x})
\label{eq:mahalanobis}
\end{equation}

In summary, our vehicle detection module combines YOLOv10 with DeepSort to achieve efficient detection and accurate tracking of vehicles in complex driving scenes. This combination not only enhances the system's detection accuracy and tracking stability but also ensures high computational efficiency, thus meeting the reliability requirements of the tail light detection pipeline.

\subsection{Tail Light State Classifier}

In the tail light state classification module, there are two components: two parallel tail light classifiers and a subsequent temporal post-processing module.

Firstly, the tracking sequences of vehicles obtained from the first module are input in parallel to the classifiers trained on our dataset. Our classifier uses ResNet34 as the backbone because it has relatively shallow layers and lower computational cost, making it suitable for applications requiring fast inference. The network is divided into four stages, each containing convolutional layers that extract features at different levels from the input data. We choose to extract features from the fourth stage of the network to leverage higher-level feature information for classification. The neck uses a Global Average Pooling layer to average the spatial dimensions (width and height) of each channel in the feature map into a single value, producing a one-dimensional feature vector. This method effectively reduces the feature dimension while retaining global contextual information, allowing the model to make more accurate decisions in classification tasks. Finally, the Linear Classification Head maps the processed feature vector to category labels, yielding classification results for brake lights or turn signals in each frame.

\begin{figure}[h!]  % 'h!' 表示尽可能将图片放置在当前位置
    \centering
    \adjustbox{max width=\textwidth}{\includegraphics[width=1\linewidth]{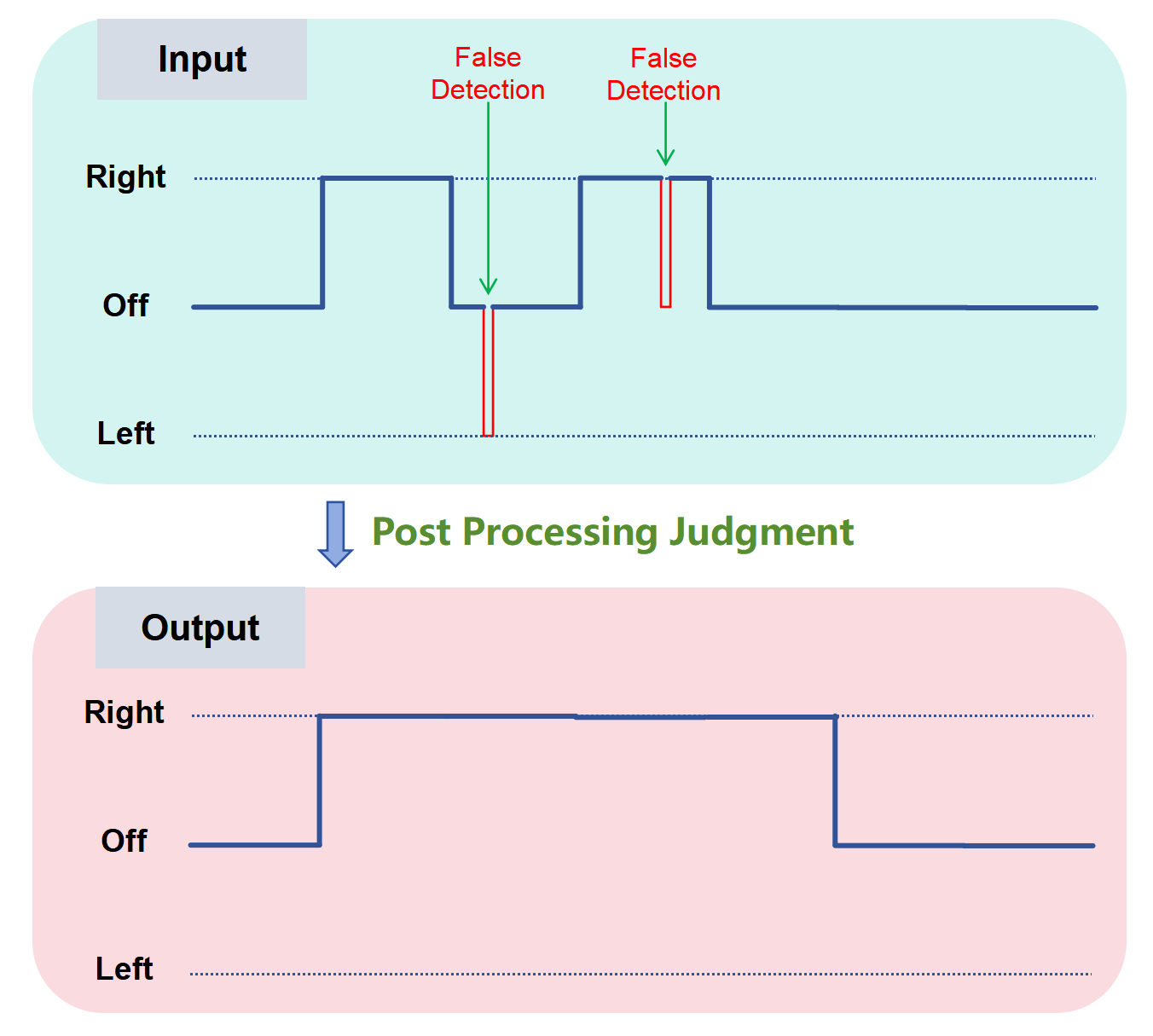}}  % 使用 adjustbox 进行自动缩放
    \caption{The introduced temporal post-processing module can filter out false detections on a single frame and output the turn signal status over a continuous period by determining the turn-on and turn-off thresholds.}
    \label{fig:pipeline3}  % 设置标签以便在文中引用
\end{figure}

At this stage, the classification results alone are not sufficient as output. In our detection system, each frame of the vehicle sequence processed by the turn signal classifier yields discrete turn signal state labels, such as "left" or "off." This means that when the left or right turn signals flash, the off state during the off moment will be classified as "off" by the single-frame classifier, leading to an alternating output of turn signal on and off. In reality, this off state should be categorized as either "left" or "right," and the discrete single-frame state is not ideal for downstream tasks that require understanding vehicle driving intentions. Additionally, potential misclassified results can introduce noise, affecting the overall judgment of the tail light state, as shown in Figure \ref{fig:pipeline3}. Therefore, an effective post-processing module is needed to consolidate these discrete results into continuous states for more accurate turn signal status determination.

% 插入图片

The proposed temporal post-processing module evaluates the discrete classification results from the classifier based on a time threshold. It then outputs the continuous tail light status over time. This module is essential for improving the reliability and accuracy of the system because it prevents single-frame detection errors from impacting the final output. When dealing with brake lights, which have simpler state transitions, a continuous state can be represented by maintaining the same state for a period of time. However, for turn signals that flash on and off, we need to capture both states. Typically, most vehicles have turn signal flash frequencies of 1-2 Hz. To accommodate this, we set activation and deactivation thresholds for turn signals. These thresholds are adjustable based on actual conditions.For example, if a detector samples at 20 Hz and a turn signal is flashing at 1 Hz, each flashing cycle lasts approximately 0.5 seconds. We set the activation threshold to 0.1 seconds. This means that if the turn signal "left" state lasts more than 0.1 seconds(equivalent to 2 frames), it is considered active. This threshold allows us to capture instantaneous changes in turn signal status while minimizing the impact of brief noise.To determine the deactivation status of the turn signal, we set a longer time threshold. If the turn signal "off" state lasts more than 0.6 seconds (i.e., 12 frames), it is considered deactivated. This threshold helps filter out brief state changes and prevents misjudgment.By using the time threshold post-processing module described above, we can output the continuous status of the tail light over time and improve the system's robustness to a certain extent.

%%%%%%%%%%%%%%%%%%%%%%%%%%%%%%%%%%%%%%%%%%%第四章%%%%%%%%%%%%%%%%%%%%%%%%%%%%%%%%%%%%%%%%%%%%

\section{EXPERIMENTS}
In this section, we will first introduce our new dataset, TLD, and then evaluate the performance of our algorithm on this dataset, including both the vehicle detection module and the tail light classification module.

\subsection{Dataset Details}

% 插入图片
\begin{figure}[h!]  % 'h!' 表示尽可能将图片放置在当前位置
    \centering
    \adjustbox{max width=\textwidth}{\includegraphics[width=1\linewidth]{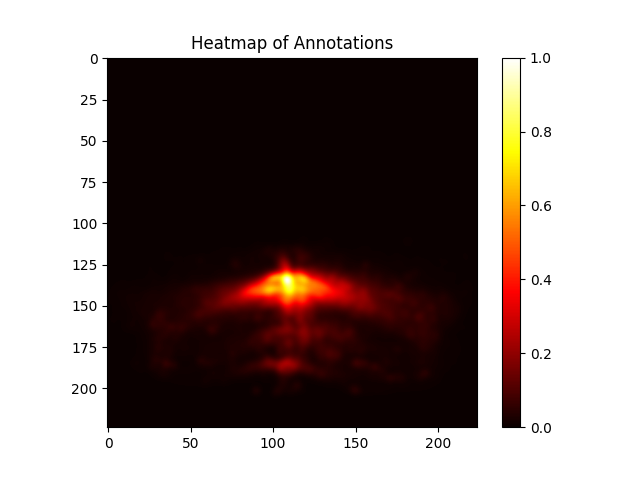}}  % 使用 adjustbox 进行自动缩放
    \caption{Heatmap of Annotation Distribution in TLD}
    \label{fig:heatmap}  % 设置标签以便在文中引用
\end{figure}

In this section, we will first introduce our new dataset, TLD, and then evaluate the performance of our algorithm on this dataset. This evaluation includes both the vehicle detection module and the tail light classification module.

Based on the video sources and the richness of the annotations, we have divided the dataset into two subsets: TLD-YT and TLD-LOKI. TLD-YT includes data from YouTube videos, while TLD-LOKI consists of annotations from the LOKI dataset.

For the YouTube video subset, we first preprocessed these driving videos by removing the first 90 seconds and the last 30 seconds. This was done to filter out unrelated content such as intros and subscription prompts. All driving videos were then annotated at a frequency of 2 Hz, resulting in a total of 111,800 frames. Each image was manually annotated for both brake light and turn signal states. In TLD-YT, we performed decoupled annotations for brake lights and turn signals. This was done to ensure that the vehicle's lighting information is comprehensive and realistic. Additionally, since tail light information is just one of the vehicle's status attributes, all tail light annotations are mapped back to their respective vehicles. Each tail light is matched to its associated vehicle in order to accurately predict the vehicle's subsequent actions. Therefore, we did not annotate the specific locations of the tail lights, but rather annotated the 2D bounding boxes of the vehicles in the real scenes. The brake light states were manually determined for each vehicle and assigned as the label for that object. Subsequently, we annotated the turn signal states with attributes including off, left, right, both, and unknown. It is worth noting that most existing car tail light datasets do not include annotations for hazard lights (both left and right turn signals on simultaneously). We consider hazard lights an important aspect of road safety, and thus annotated these cases in our dataset. To our knowledge, we are the first to include hazard light annotations in a tail light detection dataset.

In TLD-LOKI, we first filtered the Vehicle objects in the original LOKI dataset. We retained only those that were not severely occluded and visible in the field of view. We then updated the labels to four categories: off, brake, left, and right. This resulted in 40,890 annotated frames, 67,253 instances, and 644 scenarios. With the existing vehicle intention labels and additional sensor information from the LOKI dataset, we believe TLD-LOKI will provide significant value for future research.

\subsection{Experiment for Taillight Classification}

\begin{table*}[]
\centering
\caption{Experiment for Brake Signal Classification}
\label{tab:brake classification}
\begin{tabular}{c|ccccccccc}
\hline
\textbf{Methods} &
  \textbf{Classes} &
  \textbf{Precision} &
  \textbf{Recall} &
  \textbf{F1 Score} &
  \textbf{\begin{tabular}[c]{@{}c@{}}Average\\ Precision\end{tabular}} &
  \textbf{\begin{tabular}[c]{@{}c@{}}Top-1\\ Acc\end{tabular}} &
  \textbf{\begin{tabular}[c]{@{}c@{}}Mean\\ Precision\end{tabular}} &
  \textbf{\begin{tabular}[c]{@{}c@{}}Mean\\ Recall\end{tabular}} &
  \textbf{\begin{tabular}[c]{@{}c@{}}Mean\\ F1 Score\end{tabular}} \\ \hline
\multirow{2}{*}{MobileNetV3-small} &
  car\_BrakeOff &
  97.52 &
  97.84 &
  97.68 &
  99.33 &
  \multirow{2}{*}{96.65} &
  \multirow{2}{*}{95.96} &
  \multirow{2}{*}{95.73} &
  \multirow{2}{*}{95.84} \\
 &
  car\_BrakeOn &
  94.4 &
  93.62 &
  94.01 &
  97.84 &
   &
   &
   &
   \\ \hline
\multirow{2}{*}{MobileNetV3-large} &
  car\_BrakeOff &
  96.06 &
  98.28 &
  97.15 &
  99.19 &
  \multirow{2}{*}{95.86} &
  \multirow{2}{*}{95.68} &
  \multirow{2}{*}{93.97} &
  \multirow{2}{*}{94.77} \\
 &
  car\_BrakeOn &
  95.3 &
  89.66 &
  92.39 &
  97.1 &
   &
   &
   &
   \\ \hline
\multirow{2}{*}{EfficientNetV2\_b0} &
  car\_BrakeOff &
  97.68 &
  98.63 &
  98.15 &
  99.4 &
  \multirow{2}{*}{97.33} &
  \multirow{2}{*}{\textbf{97.03}} &
  \multirow{2}{*}{96.31} &
  \multirow{2}{*}{96.66} \\
 &
  car\_BrakeOn &
  96.39 &
  94 &
  95.18 &
  98.2 &
   &
   &
   &
   \\ \hline
\multirow{2}{*}{EfficientNetV2-s} &
  car\_BrakeOff &
  97.87 &
  98.53 &
  98.2 &
  99.47 &
  \multirow{2}{*}{97.4} &
  \multirow{2}{*}{97.02} &
  \multirow{2}{*}{96.52} &
  \multirow{2}{*}{96.76} \\
 &
  car\_BrakeOn &
  96.16 &
  94.49 &
  95.32 &
  98.38 &
   &
   &
   &
   \\ \hline
\multirow{2}{*}{MobileViT-s} &
  car\_BrakeOff &
  97.89 &
  98.19 &
  98.04 &
  99.51 &
  \multirow{2}{*}{97.18} &
  \multirow{2}{*}{96.61} &
  \multirow{2}{*}{96.38} &
  \multirow{2}{*}{96.49} \\
 &
  car\_BrakeOn &
  95.33 &
  94.57 &
  94.95 &
  98.25 &
   &
   &
   &
   \\ \hline
\multirow{2}{*}{Resnet34} &
  car\_BrakeOff &
  98.1 &
  98.37 &
  98.23 &
  99.42 &
  \multirow{2}{*}{\textbf{97.45}} &
  \multirow{2}{*}{96.94} &
  \multirow{2}{*}{\textbf{96.74}} &
  \multirow{2}{*}{\textbf{96.84}} \\
 &
  car\_BrakeOn &
  95.78 &
  95.12 &
  95.45 &
  98.24 &
   &
   &
   &
   \\ \hline
\multirow{2}{*}{Resnet50} &
  car\_BrakeOff &
  97.83 &
  98.53 &
  98.18 &
  99.41 &
  \multirow{2}{*}{97.37} &
  \multirow{2}{*}{96.99} &
  \multirow{2}{*}{96.46} &
  \multirow{2}{*}{96.72} \\
 &
  car\_BrakeOn &
  96.15 &
  94.4 &
  95.27 &
  98.27 &
   &
   &
   &
   \\ \hline
\multirow{2}{*}{Resnet101} &
  car\_BrakeOff &
  97.58 &
  98.37 &
  97.97 &
  99.36 &
  \multirow{2}{*}{97.07} &
  \multirow{2}{*}{96.65} &
  \multirow{2}{*}{96.05} &
  \multirow{2}{*}{96.35} \\
 &
  car\_BrakeOn &
  95.72 &
  93.74 &
  94.72 &
  98.02 &
   &
   &
   &
   \\ \hline
\end{tabular}

\end{table*}

\begin{table*}[]
\centering
\caption{Experiment for Turning Signal Classification}
\label{tab:turn classification}

\begin{tabular}{c|ccccccccc}
\hline
\textbf{Methods} &
  \textbf{Classes} &
  \textbf{Precision} &
  \textbf{Recall} &
  \textbf{F1 Score} &
  \textbf{\begin{tabular}[c]{@{}c@{}}Average\\ Precision\end{tabular}} &
  \textbf{\begin{tabular}[c]{@{}c@{}}Top-1\\ Acc\end{tabular}} &
  \textbf{\begin{tabular}[c]{@{}c@{}}Mean\\ Precision\end{tabular}} &
  \textbf{\begin{tabular}[c]{@{}c@{}}Mean\\ Recall\end{tabular}} &
  \textbf{\begin{tabular}[c]{@{}c@{}}Mean\\ F1 Score\end{tabular}} \\ \hline
\multirow{4}{*}{MobileNetV3-small} &
  off &
  98.76 &
  99.47 &
  99.11 &
  99.81 &
  \multirow{4}{*}{98.2} &
  \multirow{4}{*}{88.34} &
  \multirow{4}{*}{77.96} &
  \multirow{4}{*}{82.59} \\
 &
  right &
  84.44 &
  75.06 &
  79.48 &
  85.61 &
   &
   &
   &
   \\
 &
  left &
  86.82 &
  75.07 &
  80.52 &
  85.1 &
   &
   &
   &
   \\
 &
  both &
  83.33 &
  62.22 &
  71.25 &
  71.08 &
   &
   &
   &
   \\ \hline
\multirow{4}{*}{MobileNetV3-large} &
  off &
  95.11 &
  100 &
  97.49 &
  95.11 &
  \multirow{4}{*}{95.11} &
  \multirow{4}{*}{23.78} &
  \multirow{4}{*}{25} &
  \multirow{4}{*}{24.37} \\
 &
  right &
  0 &
  0 &
  0 &
  1.92 &
   &
   &
   &
   \\
 &
  left &
  0 &
  0 &
  0 &
  2.44 &
   &
   &
   &
   \\
 &
  both &
  0 &
  0 &
  0 &
  0.53 &
   &
   &
   &
   \\ \hline
\multirow{4}{*}{EfficientNetV2-b0} &
  off &
  98.95 &
  99.45 &
  99.2 &
  99.8 &
  \multirow{4}{*}{98.4} &
  \multirow{4}{*}{87.94} &
  \multirow{4}{*}{82.43} &
  \multirow{4}{*}{85.05} \\
 &
  right &
  87.69 &
  77.41 &
  82.23 &
  88.08 &
   &
   &
   &
   \\
 &
  left &
  87.29 &
  79.55 &
  83.24 &
  87.59 &
   &
   &
   &
   \\
 &
  both &
  77.83 &
  73.33 &
  75.51 &
  74.1 &
   &
   &
   &
   \\ \hline
\multirow{4}{*}{EfficientNetV2-s} &
  off &
  98.95 &
  99.4 &
  99.18 &
  99.85 &
  \multirow{4}{*}{98.36} &
  \multirow{4}{*}{88.24} &
  \multirow{4}{*}{80.93} &
  \multirow{4}{*}{84.19} \\
 &
  right &
  83.23 &
  83.95 &
  83.59 &
  89.34 &
   &
   &
   &
   \\
 &
  left &
  89.05 &
  76.83 &
  82.49 &
  89.3 &
   &
   &
   &
   \\
 &
  both &
  81.71 &
  63.56 &
  71.5 &
  77.28 &
   &
   &
   &
   \\ \hline
\multirow{4}{*}{MobileViT-s} &
  off &
  98.84 &
  99.07 &
  98.95 &
  99.77 &
  \multirow{4}{*}{97.9} &
  \multirow{4}{*}{85.06} &
  \multirow{4}{*}{77.98} &
  \multirow{4}{*}{80.94} \\
 &
  right &
  83.96 &
  74.94 &
  79.19 &
  84.58 &
   &
   &
   &
   \\
 &
  left &
  74.95 &
  79.26 &
  77.05 &
  81.92 &
   &
   &
   &
   \\
 &
  both &
  82.5 &
  58.67 &
  68.57 &
  62.29 &
   &
   &
   &
   \\ \hline
\multirow{4}{*}{Resnet34} &
  off &
  99.02 &
  99.57 &
  99.3 &
  99.86 &
  \multirow{4}{*}{\textbf{98.63}} &
  \multirow{4}{*}{\textbf{91.69}} &
  \multirow{4}{*}{\textbf{82.75}} &
  \multirow{4}{*}{\textbf{86.82}} \\
 &
  right &
  91.06 &
  81.73 &
  86.14 &
  91.94 &
   &
   &
   &
   \\
 &
  left &
  89.73 &
  81.69 &
  85.52 &
  90.6 &
   &
   &
   &
   \\
 &
  both &
  86.93 &
  68 &
  76.31 &
  77.9 &
   &
   &
   &
   \\ \hline
\multirow{4}{*}{Resnet50} &
  off &
  98.93 &
  99.54 &
  99.23 &
  99.79 &
  \multirow{4}{*}{98.48} &
  \multirow{4}{*}{90.23} &
  \multirow{4}{*}{81.37} &
  \multirow{4}{*}{85.44} \\
 &
  right &
  89.9 &
  80.25 &
  84.8 &
  91.11 &
   &
   &
   &
   \\
 &
  left &
  88.68 &
  78.58 &
  83.32 &
  88.26 &
   &
   &
   &
   \\
 &
  both &
  83.43 &
  67.11 &
  74.38 &
  76.12 &
   &
   &
   &
   \\ \hline
\multirow{4}{*}{Resnet101} &
  off &
  98.62 &
  99.43 &
  99.02 &
  99.81 &
  \multirow{4}{*}{97.99} &
  \multirow{4}{*}{86.21} &
  \multirow{4}{*}{75.50} &
  \multirow{4}{*}{80.29} \\
 &
  right &
  84.93 &
  70.99 &
  77.34 &
  84.19 &
   &
   &
   &
   \\
 &
  left &
  83.24 &
  71.57 &
  76.96 &
  84.15 &
   &
   &
   &
   \\
 &
  both &
  78.03 &
  60 &
  67.84 &
  70.82 &
   &
   &
   &
   \\ \hline
\end{tabular}

\end{table*}

{\bf Experimental Settings.}
We conducted experiments to evaluate the performance of tail light detection and recognition using our TLD dataset. During the training process, we started by cropping images of annotated vehicles from the labeled dataset to obtain crop images of different tail light states. These images were then organized based on their annotations. Since brake lights and turn signals were annotated separately in our dataset, we ended up with two distinct datasets: one for brake lights and one for turn signals. The brake light dataset was derived from images in TLD-YT, while the turn signal dataset was obtained from images in TLD-Full. Both datasets were randomly divided into training and testing sets in an 8:2 ratio. Data augmentation involved resizing and normalization only. To address potential overfitting caused by class imbalance in the dataset, we utilized Focal Loss with the following parameters: alpha=1, gamma=2, and loss-weight=1.0. Each method was trained for 100 epochs, and we tested various classification methods on the testing set. To comprehensively assess the performance of the classification model, especially for datasets with class imbalance or multi-label classification like the tail light dataset, we used performance metrics such as the F1 Score, which is the harmonic mean of Precision and Recall.

{\bf State-of-the-Art Methods.}We compared our two taillight state classifiers with several widely used classification network baselines, including MobileNetV3-Small/Large, EfficientNetV2-B0/S, ResNet34/50/101, and Mobile-ViT-S.

MobileNetV3 is a well-known lightweight classification network. In our experiments, we evaluated both the MobileNetV3 Small and MobileNetV3 Large versions. MobileNetV3 achieves efficient computational performance on mobile and embedded devices through hardware-aware model design and network search techniques. EfficientNetV2 is another efficient classification network that optimizes model depth, width, and resolution through a compound scaling method. This method further reduces computational resource requirements while improving efficiency and accuracy. These networks excel in lightweight and efficient performance, making them suitable for resource-constrained environments.

The ResNet series is a classic architecture in deep learning that addresses the gradient vanishing problem in deep networks by introducing residual connections. This significantly improves training performance for deep neural networks. The ResNet series includes various versions such as ResNet34, ResNet50, and ResNet101, each differing in depth. ResNet34 consists of basic residual blocks and is suited for shallower tasks. ResNet50 and ResNet101 use bottleneck residual blocks, capturing more feature information through deeper layers while maintaining computational efficiency. This enhances the model's expressiveness. Various ResNet variants have demonstrated excellent performance across different tasks.

Mobile-ViT S is a compact network based on the Vision Transformer (ViT) architecture, which combines convolutional neural networks (CNNs) and self-attention mechanisms. It consists of both CNN and ViT components. Mobile-ViT uses a lightweight CNN in the initial layers to extract low-level features, taking inspiration from MobileNet's structure, which reduces computational and parameter load. After the CNN part, Mobile-ViT incorporates a Vision Transformer module to capture global image features. The ViT processes image patches using self-attention, focusing on relationships between different positions in the image. This effectively captures long-range dependencies and achieves superior performance in classification tasks.

{\bf Experimental results.}
The experimental results for the brake light and turn signal classification tasks are presented in Tables \ref{tab:brake classification} and \ref{tab:turn classification}. We observed that ResNet networks achieved better performance in both classification tasks. Specifically, ResNet34 achieved an F1 Score of 96.84 in the brake light classification task and an F1 Score of 86.82 in the turn signal classification task. Additionally, we made the following observations:

For both the brake light and turn signal detection tasks, deeper ResNet architectures do not necessarily result in better performance. In fact, performance metrics tend to degrade to some extent. In the brake light classification task, ResNet101 exhibited a decrease in Top-1 Accuracy by 0.38, Mean Precision by 0.29, and the F1 Score dropped from 96.84 with ResNet34 to 96.35 with ResNet101, a decrease of 0.49. In the turn signal detection task, the detection accuracy is lower compared to the brake light task, and this disparity worsens with increased network depth. As shown in Table \ref{tab:turn classification}, Top-1 Accuracy decreased from 98.63 to 97.99, a drop of 0.64; Mean Precision decreased from 91.69 to 86.21, a drop of 5.48; Mean Recall decreased from 82.75 to 75.50; and Mean F1 Score decreased from 86.82 to 80.29, a drop of 6.53. The specific metrics for each category in the turn signal task in Table \ref{tab:turn classification} reveal that the main difference between ResNet34 and ResNet101 lies in the detection accuracy of left, right, and dual-flash signals. This may be due to the fact that turn signal detection involves four classification labels, and the label distribution in the dataset is imbalanced. The majority of labels belong to the "Off" category, and the imbalance in multi-class classification poses a challenge for turn signal detection. ResNet101, being deeper with more layers and parameters, has stronger learning capacity. However, with the majority of instances belonging to the "Off" category, it is more prone to overfitting and outputs more "Off" classifications, which reduces the detection accuracy for left, right, and dual-flash signals.
This phenomenon is not unique to ResNet networks; similar trends are observed with other networks listed in the table: larger networks often perform worse than smaller networks. This effect is particularly pronounced with MobileNetV3-large. As shown in the specific metrics for turn signal detection tasks (Table \ref{tab:turn classification}), MobileNetV3-large shows a score of 0 for all other categories, indicating severe overfitting to the "Off" category, with the network classifying all inputs as "Off."

%%%%%%%%%%%%%%%%%%%%%%%%%%%%%%%%%%%%%%%%%%%第五章%%%%%%%%%%%%%%%%%%%%%%%%%%%%%%%%%%%%%%%%%%%%

\section{Conclusion and Future Work}

In this work, we present the first large-scale dataset for tail light detection. The dataset includes separate annotations for brake lights and turn signals. It consists of a total of 152,690 images, which capture driving scenes in different times, weather conditions, and countries. To establish a baseline on our dataset, we designed a two-stage method. First, we track vehicles in the driving scenes. Then, we classify the tail light states in the tracked image sequences. We also implemented a post-processing step to determine the tail light states over a certain time sequence. However, our method currently lacks temporal judgment, which affects the classification accuracy. In the future, we plan to introduce frame-to-frame comparisons to determine whether the tail light has brightened or dimmed. We believe that our work represents a significant advancement towards ensuring fail-safe control of self-driving vehicles.

%%%%%%%%%%%%%%%%%%%%%%%%%%%%%%%%%%%%%%%%%%%到此结束%%%%%%%%%%%%%%%%%%%%%%%%%%%%%%%%%%%%%%%%%%%%

\vfill
\newpage
\bibliographystyle{unsrt}  
\bibliography{ref}          
\end{document}